# A Dynamic Algorithm for Longest Common Subsequence Problem using Ant Colony Optimization Technique

Arindam Chaudhuri
Lecturer (Mathematics and Computer Science),
Meghnad Saha Institute of Technology,
Nazirabad, Uchchepota, Kolkata, India
Email: arindam_chau@yahoo.co.in

**Abstract**

We present a dynamic algorithm for solving the Longest Common Subsequence Problem using Ant Colony Optimization Technique. The Ant Colony Optimization Technique has been applied to solve many problems in Optimization Theory, Machine Learning and Telecommunication Networks etc. In particular, application of this theory in NP-Hard Problems has a remarkable significance. Given two strings $A = a_1........a_n$ and $B = b_1.......b_m$ $(m \geq n)$, the traditional technique for finding Longest Common Subsequence is based on Dynamic Programming which consists of creating a recurrence relation and filling a table of size $m \times n$. The proposed algorithm draws analogy with behavior of ant colonies function and this new computational paradigm is known as Ant System. It is a viable new approach to Stochastic Combinatorial Optimization. The main characteristics of this model are positive feedback, distributed computation, and the use of constructive greedy heuristic. Positive feedback accounts for rapid discovery of good solutions, distributed computation avoids premature convergence and greedy heuristic helps find acceptable solutions in minimum number of stages. We apply the proposed methodology to Longest Common Subsequence Problem and give the simulation results. The effectiveness of this approach is demonstrated by efficient Computational Complexity. To the best of our knowledge, this is the first Ant Colony Optimization Algorithm for Longest Common Subsequence Problem.

## 1. Introduction

The problem of finding Longest Common Subsequence (LCS) [9] and its different forms of measures from a set of $n$ elements is considered as a major one in the field of Sorting and Searching in Computer Science. It is one of the most studied problems in Computer Science as it plays an important role in strings comparison. It has potential applications in many areas such as Pattern Recognition, Data Compression, Word Processing and Genetics [17]. It can be seen as a measure of closeness of two strings as it consists in finding maximum number of identical elements of two strings when preserving the order of element matters. For example, comparing two DNA sequences to see how homologous they are or in a spell checker, looking in dictionary an entry close to a misspelled word. A subsequence of a word is obtained by deleting zero or more element of word, for example *butter* is a subsequence of *butterfly*. A LCS of two words is a subsequence of both words of maximal length, for example *artica* is a LCS of *artificial* and *articulate*. This problem has two different aspects. The first one deal with various measures of presortedness and other, with the problem of generation of LCS, provided elements are arranged in an increasing sequence resulting in a longest increasing subsequence. Although first aspect has received a good deal of attention in the past, latter one is not excepting a few. This Paper deals with this second aspect of this problem. Freedman [9] has examined complexity of an algorithm that computes length $L$ of longest increasing subsequence of $S$, where $S =< x_1,............,x_n >$ is a sequence of $n$ distinct integers given by,

$$L = \max\{k : 1 \leq i_1 < i_2 <............ < i_k \leq n; x(i_1) <.......... < x(i_k)\} \quad (1)$$

with an order of $n \log_2 n$ running time and $n \log_2 n - n \log \log_2 n + O(n)$ comparison in its worst case. He showed that a substantial amount of information regarding ordering of elements in $S$ is required for value of $L$ to be determined uniquely.

In this Paper, we have concentrated on finding a low-complexity solution for LCS Problem using Ant Colony Optimization (ACO) [8] paradigm. The ACO was introduced by M. Dorigo and colleagues [8] as a novel nature-inspired metaheuristic for solution of Hard Combinatorial Optimization (CO) problems in 1990s. ACO belongs to the class of metaheuristics, which are Approximate algorithms used to obtain good enough solutions to Hard CO [6] problems in a reasonable amount of computation time. The inspiring source of ACO is foraging behavior of real ants. When searching for food, ants initially explore the area surrounding their nest in random manner. As soon as an ant finds a food source, it evaluates quantity and quality of food and carries some of it back to nest. During return trip, the ant deposits a chemical Pheromone Trail [3], [4] on ground. The quantity of Pheromone deposited, which may depend on quantity and quality of food, will guide other ants to the food source. As it has been shown in, indirect communication between the ants via Pheromone Trails enables them to find shortest paths between their nest and food sources. This characteristic of real ant colonies is exploited in artificial ant colonies in order to solve CO problems.

According to Papadimitriou and Steiglitz [16], a CO problem $P = (S, f)$ is an Optimization problem in which given a finite set of solutions $S$ called *Search Space* and an objective function given as $f : S \mapsto \Re^+$ that assigns positive cost value to each of the solutions, the goal is either to find a solution of minimum cost value, or in case of approximate solution, a good solution in a reasonable amount of time. Ant Colony Optimization algorithms belong to class of metaheuristics [5] and therefore follow latter goal. The central component of an ACO algorithm is a parameterized probabilistic model called Pheromone Model [8]. The Pheromone Model consists of a vector of model parameters $T$ called Pheromone Trail Parameters. The Pheromone Trail parameters $T_i \in T$, which are usually associated to components of solutions, have values $\tau_i$ called Pheromone values. The Pheromone Model is used to probabilistically generate solutions [3], [8] to the problem under consideration by assembling them from finite set of solution components. At runtime, ACO algorithms update Pheromone values using previously generated solutions. The update aims to concentrate search in regions of Search Space containing high quality solutions. In particular, reinforcement of solution components depending on solution quality is an important ingredient of ACO algorithms. In general, ACO approach attempts to solve an Optimization problem by repeating the two steps viz., (i) candidate solutions are constructed using Pheromone Model, i.e., parameterized probability distribution over Solution Space; (ii) the candidate solutions are used to modify Pheromone values in a way that is deemed to bias future sampling toward high quality solutions. Initially ACO was applied to Traveling Salesman Problem and later it was applied to many other Combinatorial Optimization problems. The application of ACO to Combinatorial Optimization Problem such as LCS requires definition of Constructive procedure and Local Search [1]. With respect to preceding statement, we have developed constructive algorithm called ACO-LCS in which set of Artificial Ants builds feasible solution to

LCS Problem and Local Search specialized for LCS Problem as discussed below that takes these solutions to their local optimum. The resulting algorithm is called Ant Colony Optimization with Local Search for LCS Problem. The Organization of this Paper is as follows. In section 2, we define LCS Problem and ACO technique. In next section, we present some work related to LCS Problem. In section 4, we give ACO for LCS Problem by ACO-LCS algorithm. The section 5 focuses on Stochastic Combinatorial Optimization for ACO-LCS algorithm. This formulation is followed by various Numerical Examples. The section 7 discusses complexity of ACO-LCS algorithm and its comparison with other ACO simulations. Finally, in section 8 conclusions are given.

## 2. Longest Common Subsequence

Given an alphabet $\Sigma$, an element $\Sigma^*$ is called sequence or string. A subsequence of given sequence is given sequence with some elements (possibly none) left out. Formally given a sequence $X = <x_1, \ldots, x_m>$ and another sequence $Z = <z_1, \ldots, z_k>$ is subsequence of $X$ if there exists a strictly increasing sequence $<i_1, \ldots, i_k>$ of indices of $X$ such that all $j = 1, \ldots, k$ such that $x_i = z_j$. For example, $Z = <B, C, D, B>$ is subsequence of $X = <A, B, C, B, D, A, B>$ with corresponding index sequence $<2,3,5,7>$. Given two sequences $X$ and $Y$, we say that sequence $Z$ is common subsequence of $X$ and $Y$ if $Z$ is subsequence of both $X$ and $Y$. For example, if $X = <A, B, C, B, D, A, B>$ and $Y = <B, D, C, A, B, A>$, the sequence $<B, C, A>$ is common subsequence of both $X$ and $Y$. The sequence $<B, C, A>$ is not LCS of $X$ and $Y$, however since it has length 3 and sequence $<B, C, B, A>$ which is also common to both $X$ and $Y$, has length 4. The sequence $<B, C, B, A>$ is LCS of $X$ and $Y$ as is sequence $<B, D, A, B>$, since there is no common subsequence of length 5 or greater. In LCS problem [15], we are given two sequences $X = <x_1, \ldots, x_m>$, $Y = <y_1, \ldots, y_n>$ and we wish to find maximum length common subsequence of $X$ and $Y$, provided elements are arranged in an increasing sequence. The LCS Problem has been solved using Dynamic Programming [7]. The LCS Problem has an optimal-substructure property, which is given by the following Theorem. The sub problems correspond to pairs of *prefixes* of the two input sequences. To be precise, given a sequence $X = <x_1, \ldots, x_m>$, we define $i^{th}$ prefix of $X$ for $i = 0, \ldots, m$ as $X_i = <x_1, \ldots, x_i>$.

*Theorem*: Let $X = <x_1, \ldots, x_m>$ and $Y = <y_1, \ldots, y_n>$ be sequences and let $Z = <z_1, \ldots, z_k>$ any LCS of $X$ and $Y$.

1. If $x_m = y_n$, then $z_k = x_m = y_n$ and $Z_{k-1}$ is an LCS of $X_{m-1}$ and $Y_{n-1}$.
2. If $x_m \neq y_n$, then $z_k \neq x_m$ implies that $Z$ is an LCS of $X_{m-1}$ and $Y$.
3. If $x_m \neq y_n$, then $z_k \neq y_n$ implies that $Z$ is an LCS of $X$ and $Y_{n-1}$.

The characterization of above Theorem shows that LCS of two sequences contains within it an LCS of prefixes of two sequences. Thus, LCS Problem has an optimal-substructure property [10], [11]. The Problem can easily represented recursively and a recursive solution to the problem has overlapping-sub problems property.

## 3. Ant Colony Optimization

Ant Colony Optimization algorithms are Stochastic Search procedures [18]. Their central component is Pheromone Model used to probabilistically sample search space. The Pheromone Model can be derived from *model* of tackled CO problem defined as follows.

*Definition*: A model $P = (S, \Omega, f)$ of CO problem consists of [6], [22]:

1. A *search* (*or solution*) *space* $S$ defined over finite set of discrete decision variables and set $\Omega$ of *constraints* among variables;
2. An *objective function* $f : S \rightarrow \Re^+$ to be minimized

The search space $S$ is defined as follows. Given set of $n$ *discrete variables* $X_i$ with values $v_i^j \in D_i = \{v_i^1, \ldots\ldots\ldots, v_i^{|D_i|}\}, i = 1, \ldots\ldots\ldots, n$. A variable instantiation, i.e., assignment of value $v_i^j$ to variable $X_i$ is denoted by $X_i = v_i^j$. A feasible solution $s \in S$ is complete assignment that satisfies the constraints. If set of constraints $\Omega$ is empty, then each decision variable can take any value from its domain independently of the values of other decision variables. In this case $P$ is called an *Unconstrained* Problem Model, otherwise *Constrained* Problem Model. A feasible solution $s^* \in S$ is called *Globally Optimal Solution*, [1], [5] if $f(s^*) \leq f(s), \forall s \in S$. The set of Globally Optimal Solutions is denoted by $S^* \subseteq S$. To solve a CO problem it is required to find a solution $s^* \in S^*$.

A model of CO problem [1], [5] under consideration implies finite set of solution components and pheromone model as follows. First, we call the combination of decision variable $X_i$ and one of its domain values $v_i^j$, a *solution component* denoted by $c_i^j$. Then, Pheromone Model consists of *pheromone trail parameter* $T_i^j$ for each solution component $c_i^j$. The set of all solution components is denoted by $\mathfrak{C}$. The value of Pheromone Trail parameter $T_i^j$ called *pheromone value* is denoted by $\tau_i^j$. The vector of all Pheromone Trail parameters is denoted by $T$. As a CO problem can be modeled in different ways, different models of CO problem can be used to define different Pheromone Models.

### 3.1 The framework of a basic ACO algorithm

The ACO is introduced in a form that covers all algorithms that were theoretically studied, but that is not as general as definition of ACO metaheuristic. The following algorithm captures the framework of a basic ACO algorithm. It works as follows. At each iteration, $n_a$ ants probabilistically construct solutions to CO problem under consideration, exploiting given Pheromone Model. Then, optionally Local Search procedure is applied to constructed solutions. Finally, before next iteration starts, some of the solutions are used for performing Pheromone update. This framework is explained with more details in the following [8], [14].

*InitializePheromoneValues*($T$): At start of algorithm pheromone values are all initialized to constant value $c > 0$.

*ConstructSolution(T)* : The basic ingredient of any ACO algorithm is constructive heuristic for probabilistically constructing solutions. A constructive heuristic assembles solutions as sequences of elements from finite set of solution components $\mathfrak{C}$. A solution construction starts with an empty partial solution $s^p = <>$. Then, at each construction step current partial solution $s^p$ is extended by adding feasible solution component from set $\mathfrak{R}(s^p) \subseteq \mathfrak{C} \setminus \{s^p\}$. This set is determined at each construction step by solution construction mechanism in such a way that problem constraints are met. The process of constructing solutions can be regarded as walk or path on *Construction Graph* $G_C = (\mathfrak{C}, \mathfrak{L})$ which is fully connected graph whose vertices are solution components in $\mathfrak{C}$ and whose edges are elements of $\mathfrak{L}$. The allowed walks on $G_C$ are implicitly defined by solution construction mechanism that defines set $\mathfrak{R}(s^p)$ with respect to partial solution $s^p$. The choice of solution component $c_i^j \in \mathfrak{R}(s^p)$ at each construction step is done probabilistically with respect to pheromone model. The probability for choice of $c_i^j$ is proportional to $[\tau_i^j]^\alpha [\eta(c_i^j)]^\beta$, where $\eta$ is function that assigns to each valid solution component depending on current construction step, a heuristic value which is also called *Heuristic Information*. The value of parameters $\alpha$ and $\beta$, $\alpha > 0$, $\beta > 0$ determines relative importance of pheromone value and heuristic information. Heuristic information is optional but often needed for achieving high algorithm performance. In most ACO algorithms probabilities for choosing next solution component called *Transition Probabilities* are defined as follows [8]:

$$p(c_i^j / s^p) = \{[\tau_i^j]^\alpha [\eta(c_i^j)]^\beta\} / \{\sum_{c_k^l \in \mathfrak{R}(s^p)} [\tau_i^j]^\alpha [\eta(c_k^l)]^\beta\}, \forall c_i^j \in \mathfrak{R}(s^p) \quad (2)$$

**Algorithm:** The framework of a basic ACO algorithm [8]

**input:** An instance $P$ of combinatorial optimization problem model $P = (S, \Omega, f)$
    *InitializePheromoneValues(T)*
    $s_{bs} \leftarrow NULL$
    **while** termination conditions not met **do**
        $\mathfrak{S}_{iter} \leftarrow \emptyset$
        **for** $j = 1, \ldots, n_a$ **do**
            $s \leftarrow ConstructSolution(T)$
            **if** $s$ is a valid solution **then**
                $s \leftarrow LocalSearch(s)$ {optional}
                **if** ($f(s) < f(s_{bs})$) or ($s_{bs}$ = NULL) **then** $s_{bs} \leftarrow s$
                $\mathfrak{S}_{iter} \leftarrow \mathfrak{S}_{iter} \cup \{s\}$
            **end if**
        **end for**
        $ApplyPheromone(T, \mathfrak{S}_{iter}, s_{bs})$
    **end while**
    **output:** Best-so-far solution $s_{bs}$

*LocalSearch(s)* : Local search procedure is applied for improving solutions constructed by ants. The use of such procedure is optional, though experimentally it has been observed that if available its use improves algorithm's overall performance [2], [14].

$ApplyPheromone(T, \varepsilon_{iter}, s_{bs})$ : The aim of pheromone value update rule is to increase pheromone values on solution components that have been found in high quality solutions. Most ACO algorithms use variation of following update rule [8]:

$$\tau_i^j \leftarrow (1-\rho)\tau_i^j + (\rho/\varepsilon_{upd}) \sum_{\{s \in \varepsilon_{upd} / c_i^j \in s\}} F(s)) \, for \, i = 1,\ldots,n; j = 1,\ldots,|D_i| \quad (3)$$

Instantiations of this update rule are obtained by different specifications of $\varepsilon_{upd}$ which in all cases is subset of $\varepsilon_{iter} \cup \{s_{bs}\}$, where $\varepsilon_{iter}$ is set of solutions that were constructed in current iteration and $s_{bs}$ is best-so-far solution. The parameter $\rho \in (0,1]$ is called *evaporation rate*. It has the function of uniformly decreasing all pheromone values. From practical point of view, pheromone evaporation is needed to avoid too rapid convergence of algorithm towards sub-optimal region. It implements useful form of *forgetting*, favoring exploration of new areas in search space $F : \varepsilon \mapsto \Re^+$ is function such that $f(s) < f(s') \Rightarrow +\infty > F(s) \geq F(s'), \forall s \neq s' \in \varepsilon$, where $\varepsilon$ is set of all sequences of solution components that may be constructed by ACO algorithm and that correspond to feasible solutions. $F(\bullet)$ is commonly called *quality function*.

## 4. Related Work

LCS is an NP-hard problem with applications in DNA analysis or in design of conveyor belt workstations in machine production process. A brute-force approach to solving LCS problem is to enumerate all subsequences of $X$ and check each subsequence to see if it is also subsequence of $Y$, keeping track of longest subsequence found. Each subsequence of $X$ corresponds to subset of indices $\{1,\ldots,m\}$ of $X$. There are $2^m$ subsequences of $X$, so this approach requires exponential time making it impractical for long sequences. The $O(mn)$ time algorithm for LCS Problem seems to be folk algorithm. Knuth [11] posed the question of whether sub quadratic algorithms for LCS exist. Masek and Paterson [20] answered this question in affirmative by giving an algorithm that runs in $O(mn/\log n)$ time, where $n \leq m$ and sequences are drawn from set of bounded size. For special case in which no element appears more than once in an input sequence, Szymanski [19] shows that the problem can be solved in $O((n+m)\log(n+m))$ time. Many of these results extend to the problem of computing string edit distances [21]. Many sequential algorithms for LCS Problem have been proposed in the literature and time complexity has been shown to be $\Omega(mn)$ when alphabet is not fixed and lengths of two strings are $m$ and $n$ $(m \geq n)$. Parallel algorithms have also been devised on different models. On theoretical CREW PRAM, Liu and Lin [13] proposed the fastest algorithm that requires $mn/(\log m)$ processors and takes $O(\log m)$ time. In order to offer practical solutions, researchers provided solutions on systolic model which is more realistic. When symbols are input sequentially on $n$ processor linear systolic array, tight lower bound has been achieved by Lin and Chen [12]. Their algorithm takes $m + 2n - 1$ steps.

## 5. Ant Colony Optimization for Longest Common Subsequence Problem

Here we concentrate in developing a dynamic algorithm for solving LCS Problem using ACO technique [8], [14], [19], viz., ACO-LCS algorithm, provided elements are arranged in an increasing sequence.

Given set $L$ of strings over an alphabet $\Sigma$, LCS problem consists of finding string of maximal length that is subsequence of each string in $L$. The string $B$ is subsequence of string $A$, if $A$ can be obtained from $B$ by deleting in $B$ zero or more characters. Consider for example, set $L = \{bbbaaa, bbaaab, cbaab, cbaaa\}$ over the alphabet $\Sigma = \{a,b,c\}$. A longest subsequence of $L$ is $cbbbaaab$. The character at position $j$ in string $L_i$ is denoted by $s_{ij}$. The components of resulting construction graph are $s_{ij}$'s and graph is fully connected. The constraints enforce that true subsequence of strings in $L$ has to be built. These constraints are implicitly enforced through construction policy used by ants. Pheromone Trial $\tau_{ij}$ is associated with each component $s_{ij}$. It gives desirability of choosing character $s_{ij}$ when building the subsequence.

ACO-LCS algorithm [8], [14] works as follows. Considering sequence $S = <s_1,\ldots\ldots,s_n>$ from which LCS is to be produced. Each ant iteratively starts from an alphabet or character and builds subsequences of strings in $L$ defined over alphabet $\Sigma$, independently of each other. Thus, sequence $S$ is divided into two or more subsequences. Each ant $k$ receives copy of original set of strings $L$ and initializes its subsequence to empty string. At first construction step, ant $k$ adds to subsequence it is building a character that occurs at front of at least one string $L_i \in L$ i.e., it chooses at least one component $s_{i1}$ from sequence $S$. The choice of character to be added is based on Pheromone Trials (which are initialized to one), as well as on some additional information as explained in section 6. Once a character is added, the same character is removed from front of strings on which it occurred. Then the procedure is reapplied to modified set of strings $L'$ until all characters have been removed from all strings and set $L'$ consists of empty strings.

To describe the solution construction procedure more formally, for each ant $k$ an indicator vector is defined as $v^k = (v^k_1,\ldots\ldots,v^k_l)$ with $l = |L|$. Element $v^k_i$ of $v^k$ points to front position $v^k_i$ of string $L_i$ – the position that contains character that is candidate for inclusion in subsequence. Consider for example, set $L = \{bbbaaa, bbaaab, cbaab, cbaaa\}$ for which longest subsequence is $cbbbaaab$. In this case, vector $v^k = (2,2,3,3)$ represents situation in which first character of first and second strings, as well as first two characters of third and fourth strings is already included in subsequence. The characters that are candidates for inclusion in subsequence are therefore $a$ and $b$. In fact, $s_1(v^k_1) = s_{12} = b$, $s_2(v^k_2) = s_{22} = b$, $s_3(v^k_3) = s_{33} = a$, $s_4(v^k_4) = s_{43} = a$.

At beginning of solution construction, vector $v^k$ is initialized to $v^k = (1,\ldots\ldots,1)$. The solution construction procedure is completed once indicator vector has reached the value $v^k = (|L_1|+1,\ldots\ldots,|L_l|+1)$. As mentioned earlier, at each construction step feasible neighborhood $N^k(v^k)$, that is, set of characters can be appended to the subsequence under construction is composed of characters occurring at positions pointed by indicator vector [8]:

$$N^k(v^k) = \{x \in \Sigma : \exists i, x = s_i(v^k_i)\} \qquad (4)$$

The choice of character in $x \in N^k(v^k)$ is to append to subsequence is done according to pseudorandom proportional action choice rule of ACS given by following Equation [8]:

$$j = \begin{cases} \arg\max_{l \in N_i^k}\{\tau_{il}[\eta_{il}]^\beta\}, q \leq q_0 \\ J, otherwise \end{cases} \quad (5)$$

where, $q$ is random variable uniformly distributed in $[0,1]$, $q_0 (0 \leq q_0 \leq 1)$ is parameter and $J$ is random variable selected according to probability distribution given by following Equation [8]:

$$p_{ij}^k = \frac{[\tau_{ij}]^\alpha [\eta_{ij}]^\beta}{\sum_{l \in N_l^k}[\tau_{ij}]^\alpha [\eta_{ij}]^\beta}, j \in N_l^k \quad (6)$$

where, $\tau_{ij}$ is heuristic value that is available apriori, $\alpha$ and $\beta$ are two parameters which determine the relative influence of Pheromone Trial and heuristic information; $N_l^k$ is feasible neighborhood of ant $k$ when being at city $l$, i.e., set of cities that ant $k$ has not visited yet. Using as Pheromone Trail value the sum of Pheromone Trails of all occurrences of $x$ in $l$ strings [8], [14],

$$\sum_{i:s_i(v_i^k)=x} \tau_i(v_i^k) \quad (7)$$

Finally, indicator vector is updated, that is, for $i = 1............l$ [8], [14]:

$$v_i^k = \begin{cases} v_i^k + 1, s_i(v_i^k) = x \\ v_i^k, otherwise \end{cases} \quad (8)$$

where, $x$ is character appended to subsequence.

It is to be noted that Equation (7) gives Pheromone amount to character $x$ that is (1) higher is amount of Pheromone on components $s_{ij}$ for which it holds $s_{ij} = s_i(v_i^k) = x$; (2) higher is number of times character $x$ occurs at current front of strings in $L$ (i.e., larger the cardinality of set $\{s_{ij} : s_{ij} = s_i(v_i^k) = x\}$). This later rule (2) reflects majority merge heuristic that at each step chooses character that occurs most often at front of strings. An additional variant can also be considered that weighs each Pheromone Trail with $|L_i| - v_i^k + 1$, giving higher weight to characters occurring in strings in which many characters still need to be matched by subsequence. This later choice is inspired by $L$ − majority merge (LM) heuristic that weighs each character with length of string $L_i$ and then selects character with largest sum of weights of all its occurrences.

Considering that if sequence $S$ be divided into only two subsequences viz., $a_i$ and $b_j$, where $1 \leq i \leq \lfloor n/2 \rfloor; 1 \leq j \leq \lfloor n/2 \rfloor$ [9], [10]. Thus, each subsequence is of length $\lfloor n/2 \rfloor$ which is obtained by making pair wise comparisons between each pair of elements of $S$ by ants based on amount of Pheromone deposited subject to the condition that $a_i \leq b_j$. When $n$ is odd the last element is left to be used later. Now, pair wise comparisons are made between elements of smaller set called $a$-set (say) based on amount of Pheromone deposited by ants. This is repeated until smallest element is found. The repetition procedure is generally based on Pheromone updates and Local Search done by ants, which are explained below. The similar procedure is repeated for larger set called $b$-set (say) until largest element is found. It is to be observed here that an extra phase is required when $n$ is odd, in which case last element can be compared successively with maximum and minimum element to calculate the final maximum and minimum element respectively. The repetition procedure is again based on Pheromone updates and Local Search done by ants successively [2], [8].

*Pheromone Update*: The amount of Pheromone an ant $k$ deposits is given by [8], [14]:
$$\Delta \tau^k = g(r^k)/|s^k| \quad (9)$$
where, $r^k$ is rank of ant $k$ after ordering ants according to quality of their solutions in current iteration, $g$ is some function of rank and $s^k$ is solution built by ant $k$. Pheromones are updated as follows: First, a vector $z^k = (z_1^k,\ldots\ldots,z_l^k)$ with $l = |L|$ analogous to one used for solution construction is defined and initialized to $z^k = (1,\ldots\ldots,1)$. The elements $z_i^k$ of $z^k$ points to character in string $L_i$ that is candidate for receiving Pheromone. Then $s^k$ the subsequence built by ant $k$ is scanned from first to last position. Let $x_h$ denote $h^{th}$ character in $s^k$, $h = 1,\ldots\ldots,|s^k|$. At each step of scanning procedure, first is memorized the set $M_h^k = \{s_i(z_i^k) : s_i(z_i^k) = x_h\}$ of elements in strings belonging to $L$ that are pointed by indicator vector whose value is equal to $x_h$; it is to be noted that by construction of subsequence there is at least one such character. Let $x_h$ exists. Next, indicator vector is updated, i.e., for $i = 1,\ldots\ldots,l$ [8], [14]:
$$z_i^k = \begin{cases} z_i^k + 1, s_i(z_i^k) = x_h \\ z_i^k, otherwise \end{cases} \quad (10)$$
Once subsequence has been entirely scanned, the amount of Pheromone to be deposited $k$ on component $s_{ij} \in M_h^k$ is given by [8], [14],
$$\Delta \tau_{ij}^k = (\Delta \tau^k / |M_h^k|) \cdot \{2(|s_k| - h + 1)/(|s_k|^2 + |s_k|)\} \quad (11)$$

The left term of right-hand side of Equation (11) says that Pheromone for $h^{th}$ character in $s_k$ is distributed equally among components of strings in $L$, if $x_h$ occurred in more than one string. The right term of right-hand side of Equation (11) is scaling factor that ensures that overall sum of Pheromones deposited by ant $k$ is equal to $\Delta \tau^k$; additionally, this scaling factor ensures that earlier the character occurs in $s_k$, larger the amount of Pheromone it receives. Hence, each character of string receives an amount of Pheromone that depends on how early character was chosen in construction process of ant $k$, how good ant $k's$ solution is and number of strings from which character was chosen in same construction step. Once all Pheromone Trails are evaporated and above computations are done, the contributions of all ants to characters' Pheromone Trails are summed and added to Pheromone Trail matrix.

*Local Search*: The pair wise comparison between elements in generated subsequence is based on Local Search done by ants. Local Search is an optional component of ACO algorithms [2], [8], although it has been shown since early implementations that it can greatly improve overall performance of ACO metaheuristic [5] when Static CO problems are considered. In ACO-LCS, Local Search is applied once ants have built their initial solutions and each solution is carried to its local optimum by application of Local Search routines. Locally optimal solutions are then used to update Pheromone Trails on arcs of construction graph, according to Pheromone Update procedure discussed earlier. Some of the commonly used local search procedures are: (*i*) Edge-Exchange Heuristics; (*ii*) Path-Preserving Exchange Heuristics; (*iii*) Handling Precedence

Constraints; (*iv*) Lexicographic Search Strategy with Precedence Constraints; (*v*) Labeling Procedure.

In ACO-LCS problem, any of the search procedures can be used to obtain the local optimum. A brief discussion of local search procedures can be found in [1]. The ACO-LCS algorithm is schematically represented below.

**1.** /∗ *Initialization Phase* ∗/
    **For** each pair (v, s) $\tau(v, s) = \tau_0$ **End-For**
**2.** /∗ *First step of Iteration* ∗/
    **For** k =1 to m **do**
        Let $v_k$ be the node where agent **k** is located
        $v_k \leftarrow 0$ /∗ *All ants start from node 0* ∗/
    **End-For**
**3.** /∗ *This is the step in which agents build their subsequences. The subsequence of the agent k is stored in Subsequence$_k$ in an increasing sequence* ∗/
    **For** k =1 to m **do**
     **For** i =1 to n -1 **do**
        Starting from $v_k$ compute the set $N(v_k)$ of feasible nodes
        /∗ $N(v_k)$ *contains all nodes j still to be visited and such that all nodes that have to precede j have already been inserted in the sequence* ∗/
        Choose the next node $s_k$ according to the pseudorandom proportional action choice rule of Ant Colony System in an increasing sequence by making pair wise comparisons
        **Subsequence$_k$(i)** $\leftarrow (v_k, s_k)$
        $\tau(v_k, s_k) \leftarrow \Delta \tau_{ij}^k$ /∗ *This is equation(iv)* ∗/
        $v_k \leftarrow s_k$ /∗ *New node for agent k* ∗/
     **End-For**
    **End-For**
**4.** /∗ *In this step, the local search is applied to solutions built by each ant* ∗/
    **For** k =1 to m **do**
        **Optimized_ Subsequence$_k$** $\leftarrow$ local_opt_routine(Subsequence$_k$)
    **End-For**
**5.** /∗ *In this step Pheromone Trails are updated using Pheromone Update Rule* ∗/
    **For** k =1 to m **do**
        Compute **L$_k$** /∗ *L$_k$ is the length of the Optimized_ Subsequence$_k$* ∗/
    **End-For**
    Let **L$_{best}$** be the longest **L$_k$** from beginning and **Optimized_ Subsequence$_{best}$** the corresponding sequence
    **For** each (z, s) $\in$ **Optimized_ Subsequence$_{best}$**
       $\tau(z, s) \leftarrow$ update the values /∗ *Use Pheromone Update Rule* ∗/
    **End-For**
**6. If** (End_condition = True)
    **then** Print **L$_{best}$** and **Optimized_ Subsequence$_{best}$**
    **else** goto Step 2
  **End-if**

<div align="center">**The ACO-LCS Algorithm**</div>

## 6. Stochastic Combinatorial Optimization for the ACO-LCS Algorithm

Here we discuss Stochastic Combinatorial Optimization aspect [8], [19] for ACO-LCS algorithm. By Stochastic Optimization we refer to those CO problems for which some of the variables used to define them have stochastic nature. This could be the problem components, as defined below, which can have some probability of being part of problem or not, or values taken by some variables describing the problem, or value returned by Objective Function. An artificial ant in ACO is stochastic constructive procedure that incrementally builds solution by adding opportunely defined solution components to partial solution under construction. Thus, ACO metaheuristic can be applied to LCS Problem for which constructive heuristic can be defined. The real issue here is how to map LCS Problem to a representation that can be used by artificial ants to build solutions. In the following we give a formal characterization of representation that artificial ants use and the policy they implement. We make use of Model Based Search approach for ACO-LCS algorithm.

Considering the minimization problem $(S, \Omega, f)$ of section 3 [16], the goal of minimization problem is to find an optimal solution $s^*$, i.e., feasible solution of minimum cost. The set of all optimal solutions is denoted by $S^*$. At very general level, Model Based Search approach attempts to solve this minimization problem by repeating following two steps viz., (a) Candidate solutions that are constructed using some parameterized probabilistic model, i.e., parameterized probability distribution over solution space and (b) Candidate solutions that are used to modify the model in a way it is deemed to bias future sampling toward low cost solutions. An auxiliary memory may be used in which some important information collected during the search is stored. The memory which may store, for example, information on distribution of cost values or collection of high-quality solutions, can be later used for model update. Moreover, in some cases it may be desired to build new model at every iteration rather than to iteratively update same one. For any algorithm belonging to this general scheme, two components corresponding to two steps above need to be instantiated viz., (a) Probabilistic model that allows an efficient generation of candidate solutions and (b) An update rule for model's parameters or structure.

In the rest of this section we give a discussion of two systematic approaches within Model Based Search framework, namely Stochastic Gradient Ascent and Cross-Entropy methods [8], [18], which define second component that is update rule for the model. The main characteristics of this model are positive feedback, distributed computation and use of constructive greedy heuristic. Positive feedback accounts for rapid discovery of good solutions, distributed computation avoids premature convergence and greedy heuristic helps find acceptable solutions in minimum number of stages.

We assume that CO problem $(S, \Omega, f)$ is mapped on problem that can be characterized by following list of items [16]:

1. A finite set $C = \{c_1, \ldots\ldots\ldots\ldots, c_{N_c}\}$ of *components*, where $N_c$ is number of components.

2. A finite set $X$ of *states* of the problem, where state is sequence $x = <c_i, c_j, \ldots\ldots\ldots, c_k, \ldots\ldots\ldots>$ over elements of $C$. The length of sequence $x$ i.e., number of components in sequence is expressed by $|x|$. The maximum length of sequence is bounded by positive constant $n < +\infty$.

3. The set of (candidate) solutions $S$ is subset of $X$ (i.e., $S \subseteq X$).

4. A set of feasible states $\tilde{X}$ with $\tilde{X} \subseteq X$ defined via a set of *constraints* $\Omega$.

5. A non-empty set $S^*$ of optimal solutions with $S^* \subseteq \tilde{X}$ and $S^* \subseteq S$.

Given above formulation artificial ants build candidate solutions by performing randomized walks on completely connected weighted graph $G_C = (\mathfrak{C}, \mathfrak{L})$ which is fully connected graph whose vertices are solution components in $\mathfrak{C}$ and whose edges are elements of $\mathfrak{L}$. Further vector $T$ is considered gathering so-called *Pheromone Trails* $\tau$. The graph $G$ is called *Construction Graph*. Each artificial ant is put on randomly chosen vertex of graph and then it performs randomized walk by moving at each step from vertex to vertex in graph in such a way that next vertex is chosen stochastically according to strength of pheromone currently on arcs. While moving from one node to another of graph $G_C$, constraints $\Omega$ may be used to prevent ants from building infeasible solutions. Formally, solution construction behavior of generic ant can be described as follows [8], [19]:

**ANT_SOLUTION_CONSTRUCTION**

for each ant:
- select start node $c_1$ according to some problem dependent criterion
- set $k = 1$ and $x_k = <c_1>$

while $x_k = <c_1,...............,c_k> \in \tilde{X}$, $x_k \notin S$ and $J_{xk} \neq \Phi$ do:

  at each step $k$ after building sequence $x_k$, select next node (component)

  $c_{k+1}$ randomly using following distribution:

$$P_T(c_{k+1} = c \mid x_k) = \begin{cases} \{F(c_k,c)(\tau(c_k,c))\} / \sum_{(c_k,y) \in J_{xk}} \{F(c_k,y)(\tau(c_k,y))\}, (c_k,y) \in J_{xk} \\ 0, otherwise \end{cases} \quad (12)$$

where, connection $(c_k,y)$ belongs to $J_{xk}$ iff sequence $x_{k+1} = <c_1,...............,c_k,y>$ satisfies constraints $\Omega$ i.e., $x_{k+1} \in \tilde{X}$ and $F_{ij}(z)$ is some monotonic function most commonly, $z^\alpha \eta(i,j)^\beta$, where $\alpha, \beta > 0$ and $\eta$ are heuristic visibility values. If at some stage $x_k \notin S$ and $J_{xk} \neq \Phi$ i.e., construction process has reached a dead-end, the current state $x_k$ is discarded. For certain problems, one may find useful to use more general scheme, where $F$ depends on pheromone values of several related connections, rather than just single one. The probabilistic rule given by above equation, together with underlying Construction Graph, implicitly defines first component of Model Based Search algorithm viz., probabilistic model. Having chosen the probabilistic model, next step is to choose parameter update mechanism. In the following, we describe updates within the ant colony optimization framework as well as the ones derived from Stochastic Gradient Ascent algorithm and Cross-Entropy method [8], [19].

Many different schemes for pheromone update have been proposed within ACO framework. Most pheromone updates can be described using following generic scheme [8], [14]:

$$\forall s \in S_t', \forall (i,j) \in s : \tau(i,j) \leftarrow \tau(i,j) + Q_f(s \mid S_1,..........,S_t),$$
$$\forall (i,j) : \tau(i,j) \leftarrow (i - \rho) \cdot \tau(i,j) \quad (13)$$

where, $S_i$ is sample in $i^{th}$ iteration, $\rho, 0 < \rho < 1$ is evaporation rate and $Q_f(s \mid S_1,..........,S_t)$ is some quality function which is typically required to be non-increasing with respect to $f$ and is defined over reference set $S_t'$. Different ACO algorithms may use different quality functions and reference sets. For example, in very first ACO algorithm viz., Ant System, quality function was $1/f(s)$ and reference set $S_t' = S_t$. In more recently proposed scheme, called *iteration best update,* reference set was singleton containing best solution within $S_t$ (if there were several iteration-best solutions, one of them was chosen randomly). For *global best update*, reference set contained best among all iteration-best solutions (and if there were more than one global-best solution, earliest one was chosen).

In case good lower bound on optimal solution cost is available, one may also use the following quality function [8], [19]:

$$\overline{Q}_f(s \mid S_1,\ldots,\overline{S}_t) = \tau_0\{1 - \frac{(\bar{f}(s) - LB)}{(\bar{f} - LB)}\} = \tau_0\{\frac{(\bar{f} - f(s))}{(\bar{f} - LB)}\} \quad (14)$$

where, $\bar{f}$ is average of costs of last $k$ solutions and $LB$ is lower bound on optimal solution cost. With this quality function, solutions are evaluated by comparing their cost to average cost of other recent solutions, rather than by using absolute cost values. In addition, quality function is automatically scaled based on proximity of average cost to lower bound. Pheromone update which slightly differs from generic update described above was used in *ant colony system.* There pheromones are evaporated by ants online during solution construction, hence only pheromones involved in construction evaporate. Two additional modifications of generic update were also available. The first one uses maximum and minimum pheromone trail limits. With this modification, probability to generate any particular solution is kept above some positive threshold, which helps preventing search stagnation and premature convergence to suboptimal solutions. The second modification, proposed under the name of *hyper-cube framework* in context of combinatorial problems with binary coded solutions, is to normalize quality function, hence obtaining an automatic scaling of pheromone values $\tau_i$. While all updates described above are of somewhat heuristic nature [8], Stochastic Gradient Ascent and Cross-Entropy methods allow deriving parameters update rules in more systematic manner which are discussed below.

## 6.1 The Stochastic Gradient Ascent Update in ACO-LCS

An Update Rule for the Stochastic Gradient is [8], [19]:

$$T^{t+1} = T^t + \alpha_t \sum_{s \in S_t} Q_f(s) \nabla \ln P_T(s) \quad (15)$$

where, $S_t$ is sample at stage $t$. In case distribution is implicitly defined by ACO-type construction process, parameterized by vector of pheromone values $T$, gradient $\nabla \ln P_T(s)$ can be efficiently calculated. The following is generalized calculation.

From definition of **ANT_SOLUTION_CONSTRUCTION** [8], [14], [19], it follows that for $s = <c_1, c_2, \ldots>$,

$$P_T(s) = \prod_{k=1}^{|s|-1} P_T(c_{k+1} \mid pref_k(s)) \quad (16)$$

where, $pref_k(s)$ is $k-prefix$ of $s$ and consequently,

$$\nabla \ln P_T(s) = \prod_{k=1}^{|s|-1} \ln P_T(c_{k+1} \mid pref_k(s)) \quad (17)$$

Finally, given pair of components $(i, j) \in C^2$, using expression for $P_T(c_{k+1} = c \mid x_k)$ and assuming differentiability of $F$, it is easy to verify that if $i = c_k, j = c_{k+1}$ then,

$$\frac{\partial}{\partial_\tau (i,j)} \{\ln P_T(c_{k+1} \mid pref_k(s))\} = \frac{\partial}{\partial_\tau (i,j)} \{\ln F(\tau(i,j)) / \sum_{(i,y) \in J_x(k)} F(\tau(i,y))\}$$

$$= \{F'(\tau(i,j))/F(\tau(i,j))\} - \{F'(\tau(i,j))/\sum_{(i,y) \in J_x(k)} F(\tau(i,y))\} \quad (18)$$

$$= \{1 - F(\tau(i,j))/\sum_{(i,y) \in J_x(k)} F(\tau(i,y))\}\{F'(\tau(i,j))/F(\tau(i,j))\}$$

$$= \{1 - P_T(j \mid pref_k(s))\}\{G(\tau(i,j))\}$$

where, $G(\cdot) = F'(\cdot)/F(\cdot)$ and subscript of $F$ was omitted for clarity of presentation. If $i = c_k, j \neq c_{k+1}$ then,

$$\frac{\partial}{\partial_\tau (i,j)} \{\ln P_T(c_{k+1} \mid pref_k(s))\} = -P_T(j \mid pref_k(s))G(\tau(i,j)) \quad (19)$$

If $i \neq c_k$, then $P_T(c_{k+1} \mid pref_k(s))$ is independent of $\tau(i,j)$ and

$$\frac{\partial}{\partial_\tau (i,j)} \{\ln P_T(c_{k+1} \mid pref_k(s))\} = 0 \quad (20)$$

By combining these results, following pheromone update rule is derived:

$$\forall s \in S_t', \forall (i,j) \in s : \tau(i,j) \leftarrow \tau(i,j) + \alpha_t Q_f(s)G(\tau(i,j)),$$
$$\forall s = \{c_1, \ldots c_k, \ldots\} \in S_t, \forall i = c_k, 1 \leq k \leq |s|, \quad (21)$$
$$\forall j : \tau(i,j) \leftarrow \alpha_t Q_f(s) P_T(j \mid pref_k(s))G(\tau(i,j))$$

Hence, any connection $(i, j)$ used in construction of solution is reinforced by an amount $\alpha_t Q_f(s)G(\tau(i,j))$ and any connection considered during construction has its pheromone values evaporated by an amount $\alpha_t Q_f(s) P_T(j \mid pref_k(s))G(\tau(i,j))$. It is to be noted that, if solutions are allowed to contain loops a connection may be updated more than once for same solution. In order to guarantee stability of resulting algorithm, it is desirable to have bounded gradient $\nabla \ln P_T(s)$. This means that function $F$, for which $G = F'/F$ is bounded and should be used.

## 6.2 The Cross-Entropy Update in ACO-LCS

The Cross-Entropy approach requires solving the following intermediate problem [8], [19]:

$$P_{t+1} = \arg\max_{P \in M} \sum_{s \in S_t} Q_f(s) \ln P(s) \quad (22)$$

Let us now consider this problem in more details in case of an ACO-LCS type probabilistic model. Since at maximum the gradient must be zero [8], [19], we have:

$$\sum_{s \in S_t} Q_f(s) \ln P(s) = 0 \quad (23)$$

In some relatively simple cases, for example when solution $s$ is represented by an unconstrained string of bits of length $n$, $(s_1,\ldots\ldots,s_n)$ and there is single parameter $\tau_i$ for $i^{th}$ position in string, such that $P_T(s) = \prod_i p_\tau(s_i)$, the above Equation system reduces to set of independent Equations [8], [19]:

$$\frac{d}{d\tau}\{\frac{\ln p_\tau}{\sum_{\substack{s \in S_t \\ s_i=1}} Q_f(s)}\} = -\frac{d}{d\tau}\{\frac{\ln(1-p_\tau)}{\sum_{\substack{s \in S_t \\ s_i=0}} Q_f(s)}\}, i=1,\ldots\ldots,n \quad (24)$$

which may often be solved analytically. For example, $p_\tau = \tau$ it can be verified that solution of above Equation is [8], [19]:

$$p_\tau = \tau = \sum_{s \in S_t} Q_f(s)s_i / \sum_{s \in S_t} Q_f(s) \quad (25)$$

and in fact, a similar solution also applies to more general class of Markov chain models. Now, since Pheromone Trails $\tau_i$ in above equation are random variables, whose values depend on particular sample. To make the algorithm more robust, some conservatism is introduced into the update. For example, rather than discarding old pheromone values, new values may be taken to be convex combination of old values and solution of above Equation is [8], [19]:

$$\tau_i \leftarrow (1-\rho)\tau_i + \rho \sum_{s \in S_t} Q_f(s)s_i / \sum_{s \in S_t} Q_f(s) \quad (26)$$

The resulting update is identical to one used in hyper-cube framework for ACO. However, for many cases of interest equation $\sum_{s \in S_t} Q_f(s)\ln P(s) = 0$ are coupled and an analytical solution is unavailable. Nevertheless, in actual implementations of Cross-Entropy method update was of form given by Equation, $p_\tau = \tau$ which may be considered as an approximation to exact solution of Cross-Entropy Minimization problem.

Since, in general exact solution is not available an iterative scheme such as Gradient Ascent could be employed. As shown previously, gradient of log-probability may be calculated as follows [8], [19]:

If $i = c_k, j = c_{k+1}$ then,

$$\frac{\partial}{\partial_\tau(i,j)}\{\ln P_T(c_{k+1} | pref_k(s))\} = \{1 - P_T(j | pref_k(s))G(\tau(i,j))\} \quad (27)$$

where, $G(\cdot) = F'(\cdot)/F(\cdot)$ and subscript of $F$ was omitted for clarity of presentation. If $i = c_k, j \neq c_{k+1}$ then,

$$\frac{\partial}{\partial_\tau(i,j)}\{\ln P_T(c_{k+1} | pref_k(s))\} = -P_T(j | pref_k(s))G(\tau(i,j)) \quad (28)$$

If $i \neq c_k$, then $P_T(c_{k+1} | pref_k(s))$ is independent of $\tau(i,j)$ and

$$\frac{\partial}{\partial_\tau(i,j)}\{\ln P_T(c_{k+1} | pref_k(s))\} = 0 \quad (29)$$

and these values may be plugged into any general iterative solution scheme of Cross-Entropy Minimization problem. It may be concluded that if Equation, $p_\tau = \tau$ is used as possible approximate solution of Equation [8], [19]:

$$P_{t+1} = \arg\max_{P \in M} \sum_{s \in S_t} Q_f(s) \ln P(s) \qquad (30)$$

the same pheromone update rule as in hyper-cube framework for ACO is derived. If otherwise a single-step Gradient Ascent is used for solving the problem given by above Equation, generalization of Stochastic Gradient Ascent Update is obtained in which quality function is allowed to change over time.

## 7. Numerical Examples

In this section we consider three types of sequences namely a sequence [11] consisting of $n$ elements when $n$ is a power of 2, a sequence consisting of $n$ elements when $n$ is even and finally a sequence consisting of $n$ elements when $n$ is odd. These sequences are formed by the elements which are determined by the amount of Pheromone deposited by ants. In first case, suppose a sequence consist of 8 distinct elements and we process this sequence such that LCS can be generated. Since, $8! = 40320$, and this number of possible cases are to be studied, we reduce this number of cases initially by substantial amount by making pair wise comparisons. This is done by drawing a directed path in following way. The directed path from one element to other has Pheromone deposited by ants.

$$a_1 \text{--------------->} b_1$$
$$a_2 \text{--------------->} b_2$$
$$a_3 \text{--------------->} b_3$$
$$a_4 \text{--------------->} b_4$$

where, $a_i$'s, $1 \leq i \leq n/2$ and $b_j$'s, $1 \leq j \leq n/2$ are sets of smaller and larger elements. As $a_i$'s and $b_j$'s contains four elements each, a combination of 576 is possible. The proposed algorithm specifically selects some permutations which has higher amount of Pheromone, by deleting those permutations of $a_i$'s and $b_j$'s sets which are identical in nature. This identical nature is determined by tracing a path from minimum to maximum. As a result of this, only 8 possible cases arise in both sides and a combination of 64 is possible. The 8 possible $a_i$'s set are: {1234, 1243, 2143, 2413, 2341, 2134, 4321, 3214}. The identical nature of {1234} with respect to other permutations is shown below:

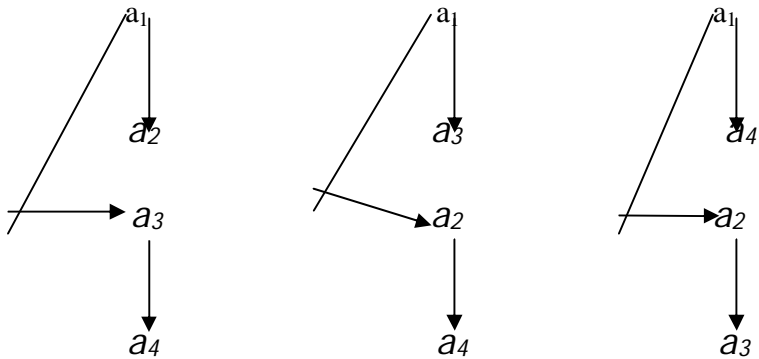

Similarly, the possible $b_j$'s sets are:

{1234, 1423, 1243, 1432, 2143, 3124, 4231, 4123}

The identical nature of {1234} with respect to other permutations is shown below:

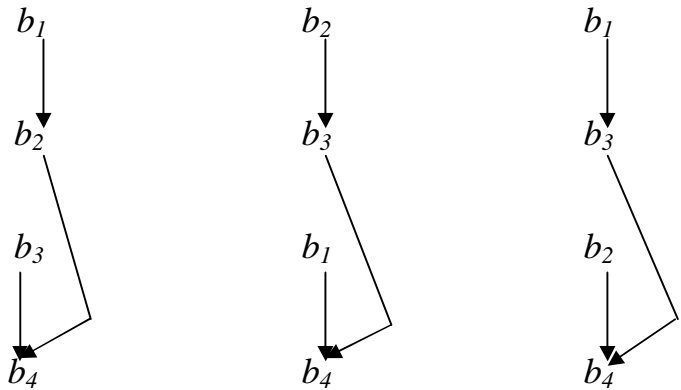

So, there are 64 combinations possible among $a_i$'s and $b_j$'s. We observe the following after some logical computational steps:

The 16 sets of length 6 are:

{$a$:1234, $b$:1432}, {$a$:1234, $b$:4231}, {$a$:1243, $b$:1423}, {$a$:1243, $b$:4123}
{$a$:2143, $b$:1423}, {$a$:2143, $b$:4123}, {$a$:2413, $b$:3124}, {$a$:2413, $b$:2143}
{$a$:2341, $b$:3124}, {$a$:2341, $b$:2143}, {$a$:2134, $b$:4231}, {$a$:2134, $b$:1423}
{$a$:3214, $b$:1234}, {$a$:3214, $b$:1243}, {$a$:4321, $b$:1234}, {$a$:4321, $b$:1243}

The typical diagram of {$a$:1234, $b$:1432} will be:

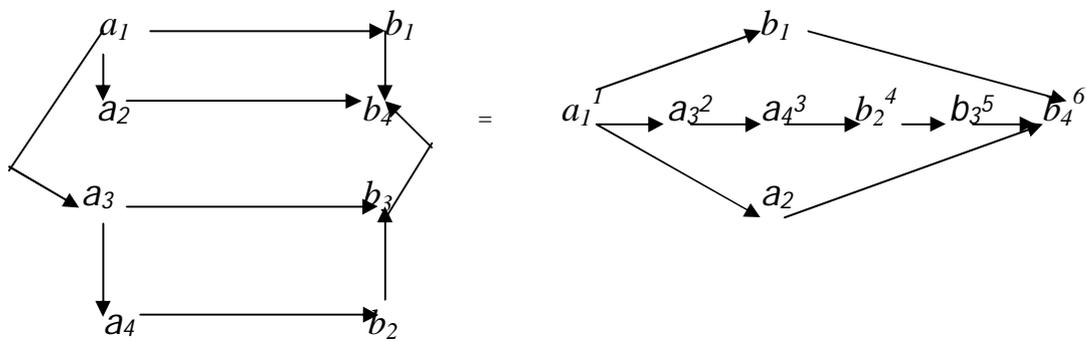

The 8 sets of length 4 will be:

{$a$:1234, $b$:1234}, {$a$:1243, $b$:1243}, {$a$:2143, $b$:2143}, {$a$:2413, $b$:1423}
{$a$:2341, $b$:1432}, {$a$:2134, $b$:3124}, {$a$:4321, $b$:4231}, {$a$:3214, $b$:4123}

The typical diagram of {$a$:1234, $b$:1234} is shown below:

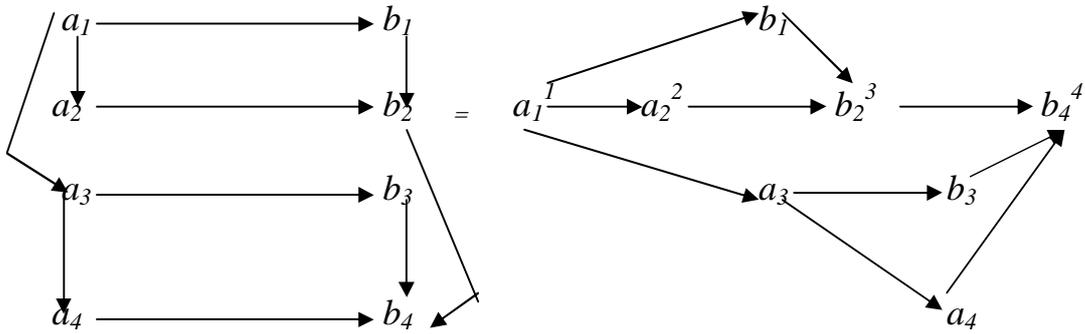

The 40 sets of length 5 will be:

{$a$:1234, $b$:1423}, {$a$:1234, $b$:1423},………………….……………….,{$a$:4321, $b$:1432}

The typical diagram of {$a$:1234, $b$:1423} is:

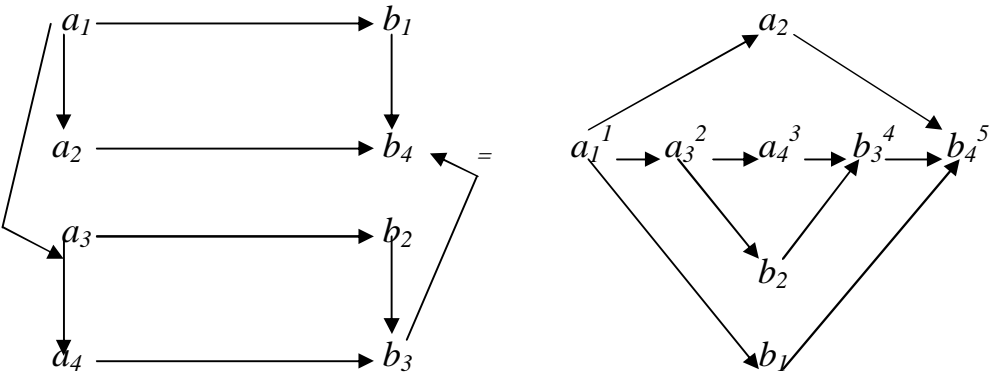

Now, we consider the second type of sequence which consists of even number of elements (say 6), then the 6 sets of length 5 each will be:

{$a$:123, $b$:213}, {$a$:213, $b$:123}, {$a$:231, $b$:123}, {$a$:231, $b$:312}, {$a$:321, $b$:132}, {$a$:321, $b$:213}

The typical diagram of {$a$:123, $b$:213} is as follows:

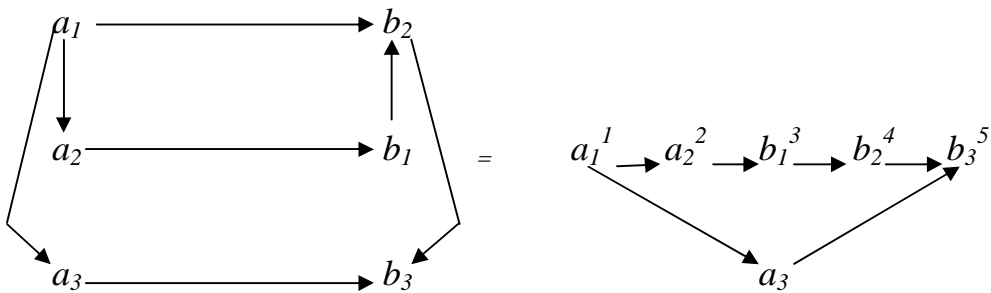

The 8 sets of length 4 will be:

{$a$:123, $b$:123}, {$a$:123, $b$:132}, {$a$:123, $b$:312}, {$a$:213, $b$:132}, {$a$:213, $b$:213}, {$a$:213, $b$:312}, {$a$:231, $b$:132}, {$a$:321, $b$:312}

The typical diagram of {$a$:123, $b$:123} is as follows:

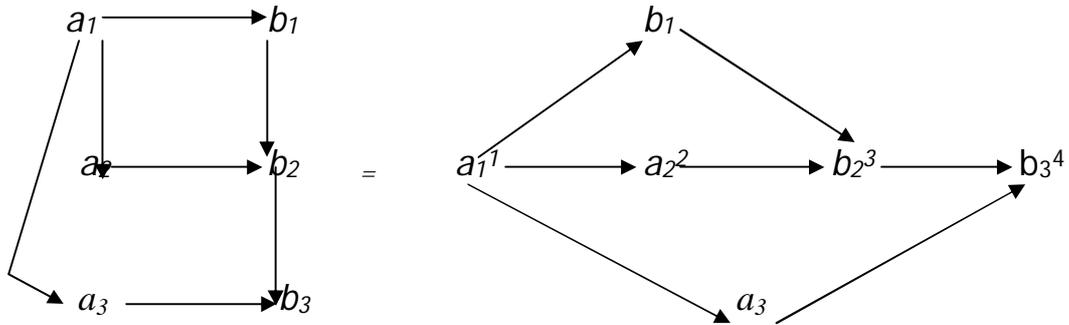

The 2 sets of length 6 is the best possible case and yields a fully sorted sequence which will be {$a$:231, $b$:213}, {$a$:321, $b$:123}. The typical diagram of {$a$:231, $b$:213} is as follows:

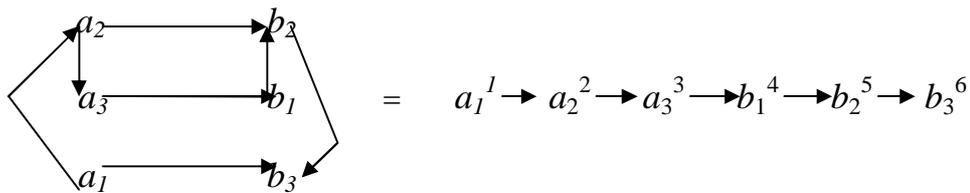

Now, we consider the final case i.e., when $n$ is odd. In this case, an additional step is required to determine both the intermediate maximum and minimum element as show below. Consider a sequence consisting of odd number of elements (say 5) which shows the 6 sets of length 4 as follows:

{$a$:12, $b$:12, 1}, {$a$:12, $b$:21, 1}, {$a$:21, $b$:12, 1}, {$a$:21, $b$:21, 1}, {$a$:12, $b$:12, 1}, {$a$:21, $b$:21, 1}, where 1 is the last element.

The typical diagram of {$a$:12, $b$:12, 1} is as follows:

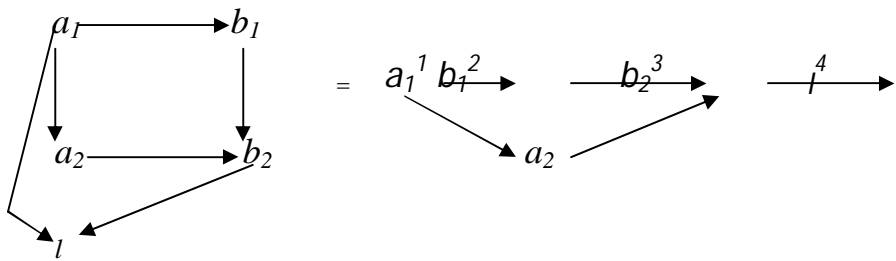

The 2 sets of length 3 will be {$a$:21, $b$:21, 1}, {$a$:12, $b$:12, 1}. The typical diagram of {$a$:21, $b$:21, 1} will be:

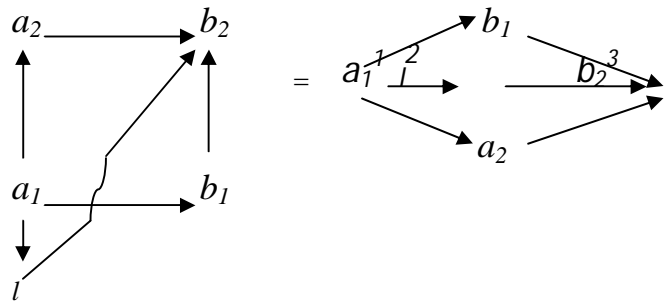

and the 4 sets of length 5 each, which is the best possible case and yields a fully sorted sequence as {$a$:12, $b$:21, 1}, {$a$:21, $b$:12, 1}, {$a$:12, $b$:21, 1}, {$a$:21, $b$:12, 1}. The typical diagram of {$a$:12, $b$:21, 1} and {$a$:21, $b$:12, 1} will be:

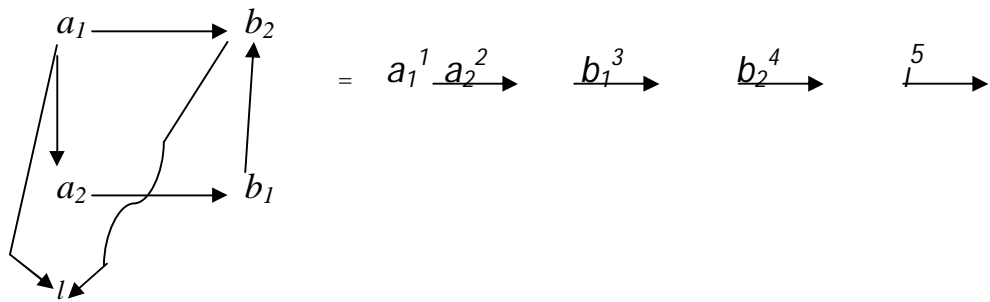

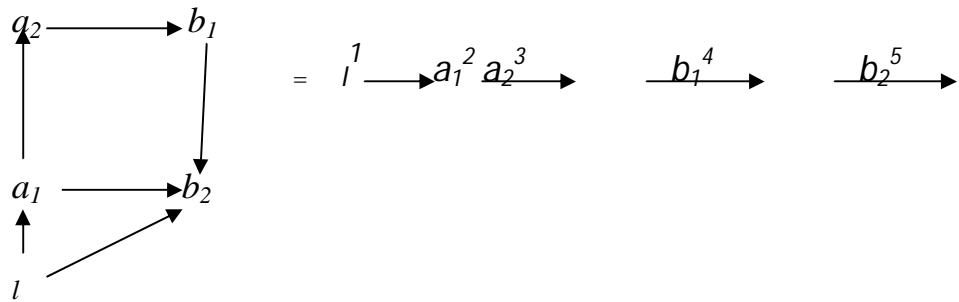

## 8. Computational Complexity and Comparison of the ACO-LCS Problem with other ACO Simulations

In this section we first present a brief discussion of the Computational Complexity of ACO-LCS Problem and then give comparison of ACO-LCS Problem with other ACO simulations [8], [14], [19]. The arguments about asymptotic worst-case time complexities of various problems have been an important issue since last few decades. To throw some light on question: *How fast can LCS be generated using ACO technique provided elements are arranged in an increasing sequence?* and its relation to central question of computational complexity theory, *Is P = NP?* As such LCS is NP-hard problem, if number of characters in sequence becomes very large value. Many algorithms in past with polynomial time-complexity have been proposed to solve LCS problem. ACO-LCS algorithm generates LCS in an increasing sequence with logarithmic complexity as given in following discussion.

Computational complexity analysis of ACO-LCS algorithm has been classified under three aspects viz., (i) time complexity (ii) comparison complexity and (iii) complexity of finding average cost. For solution of first part, the phases of algorithm have been used which uniquely determine processing time to be $\lfloor \log_2 n \rfloor$, when $n$ is power of 2 and even. But an extra phase is required when number of elements is odd which results processing time to be of $\lfloor \log_2 n \rfloor$ in its worst case. For second part i.e., comparison complexity, $\lceil 3n/2 \rceil - 2$ comparisons are required. It was found that this complexity is theoretically best possible bound. For final part of complexity analysis, average cost is determined as follows:

Initially cost is determined by number of elements that finally belongs to LCS and average cost is calculated by averaging total cost of longest, shortest and middle increasing subsequence. For Example, in case of 8 elements, set of 6 elements, longest subsequence belongs in 16 pairs of *a* and *b* and set of 4 elements, lowest subsequence belongs in 8 pairs of *a* and *b*, but set of 5 elements, which is called middle increasing subsequence, belongs in 40 pairs of *a* and *b*. So it is necessary to count all above mentioned cases for finding average cost. By observing more results in this way, it is observed that, *length of middle increasing sequence ≤ average cost ≤ length of longest increasing sequence*.

The differences between ACO-LCS with other ACO simulations are briefly summarized in three main aspects [8], [14], [19]:

a. First, ACO-LCS uses an optimal *look-ahead function* that takes into account *quality* of partial solutions (i.e., partial subsequences) that can be reached in following construction steps. To do so, ACO-LCS tentatively adds character *x* to subsequence and generates vector $\tilde{v}_k$ that is obtained by tentative addition of $x$. Then, maximum amount of pheromone on any of characters pointed by $\tilde{v}_k$ is determined. In ACO-LCS, *look-ahead function* plays role of heuristic information and is therefore indicated by $\eta$ as usual. It is defined by the expression: $\eta(x, \tilde{v}_k) = \max\{\tau_{ij}^k : s_{ij} = s_i(\tilde{v}_i^k) = x\}$. As usual with heuristic information, value $\eta(x, \tilde{v}_k)$ is weighted by an exponent $\beta$ when using pseudorandom proportional action choice rule. It is possible to extend one step look ahead to deeper levels.

b. Second, at each step of solution construction procedure, an ant although adding single character to subsequence under construction, can visit in parallel more than one component $s_{ij}$ of construction graph.

c. Finally, ACO-LCS algorithm has been implemented as parallel algorithm based on island model approach, where whole ant colony is divided into several subpopulations that occasionally exchange solutions. In this approach, two different types of colonies are considered viz., forward and backward colonies. The backward colony works on set $\acute{L}$ obtained from set *L* by reversing order of strings.

Three variants of ACO-LCS, which used (i) No *look-ahead* (ii) *look-ahead* of depth one and (iii) *look-ahead* of depth two, were compared with LM heuristic using same three levels of look ahead and with GA designed for LCS problem [8], [14], [19]. The comparison was done on three classes of instances (i) randomly generated strings (ii) strings that are similar to type of strings arising in variety of applications and (iii) several special cases, which are known to be hard for LM heuristic. The computational results showed that ACO-LCS variants, when compared to LM variants or GA for LCS performed well on instances of classes (ii) and (iii). The addition of *look-*

*ahead* proved to increase strongly solution quality for all instance classes, at cost of additional computation time. The addition of backward colonies gave substantial improvements only for special strings. Overall, ACO-LCS proved to be one of the best performing heuristics for LCS problem; this is true, in particular for structured instances which occur in real-world applications.

## 9. Conclusion

In this Paper, we considered LCS Problem and presented a dynamic algorithm using ACO technique. We apply the proposed methodology to LCS Problem and give the simulation results. The proposed algorithm viz., ACO-LCS draws analogy with behavior of ant colonies function and yields better results than traditional technique for finding the LCS based on Dynamic Programming as is evident from its efficient Computational Complexity. We also give Stochastic Combinatorial Optimization aspect for the proposed technique, which are characterized by positive feedback, distributed computation and use of constructive greedy heuristic. Positive feedback accounts for rapid discovery of good solutions, distributed computation avoids premature convergence and greedy heuristic helps find acceptable solutions in minimum number of stages. Finally, we obtained computational complexity of ACO-LCS algorithm which is $\lceil \log_2 n \rceil$ in its worst case.

**Acknowledgements**

We wish to thank all anonymous referees for their many useful comments. I dedicate this paper to the memory of my beloved father Bimal Krishna Chaudhuri.